\documentclass[letterpaper]{article} 
\usepackage{aaai2026}  
\usepackage{times}  
\usepackage{helvet}  
\usepackage{courier}  
\usepackage{xcolor}
\usepackage[hyphens]{url}  
\usepackage{graphicx} 
\usepackage{natbib}  
\usepackage{caption} 
\usepackage{pifont}
\usepackage{multirow}
\usepackage[table]{xcolor}
\usepackage{enumitem}
\definecolor{mygray}{RGB}{235,235,235}
\definecolor{mygreen}{RGB}{230,240,240}
\definecolor{myyellow}{RGB}{255,240,193}
\definecolor{gree}{RGB}{226,239,213}
\definecolor{greegray}{RGB}{245,249,241}
\usepackage{algorithm}
\usepackage{algorithmic}
\usepackage{newfloat}
\usepackage{listings}
\usepackage{amsfonts}
\DeclareCaptionStyle{ruled}{labelfont=normalfont,labelsep=colon,strut=off} 
\floatstyle{ruled}
\newfloat{listing}{tb}{lst}{}
\floatname{listing}{Listing}
\title{ICLR: Inter-Chrominance and Luminance Interaction for Natural Color Restoration in Low-Light Image Enhancement}
\author {
    Xin Xu\textsuperscript{\rm 1,2},
    Hao Liu\textsuperscript{\rm 1},
    Wei Liu\textsuperscript{\rm 1},
    Wei Wang\textsuperscript{\rm 1,2},
    Jiayi Wu\textsuperscript{\rm 1},
    Kui Jiang\textsuperscript{\rm 3}\thanks{Corresponding author.}
}
\affiliations {
    \textsuperscript{\rm 1}School of Computer Science and Technology, Wuhan University of Science and Technology, Wuhan 430065 China\\
    \textsuperscript{\rm 2}Hubei Province Key Laboratory of Intelligent Information Processing and Real-time Industrial System\\
    \textsuperscript{\rm 3}School of Computer Science and Technology, Harbin Institute of Technology, Harbin 150001 China\\
    xuxin@wust.edu.cn, liuh85872@wust.edu.cn, liuwei@wust.edu.cn, wangwei8@wust.edu.cn, wuaddone@wust.edu.cn, jiangkui@hit.edu.cn
}

\usepackage{bibentry}

\begin{document}

\maketitle

\begin{abstract}
Low-Light Image Enhancement (LLIE) task aims at improving contrast while restoring details and textures for images captured in low-light conditions. HVI color space has made significant progress in this task by enabling precise decoupling of chrominance and luminance. However, for the interaction of chrominance and luminance branches, substantial distributional differences between the two branches prevalent in natural images limit complementary feature extraction, and luminance errors are propagated to chrominance channels through the nonlinear parameter. Furthermore, for interaction between different chrominance branches, images with large homogeneous-color regions usually exhibit weak correlation between chrominance branches due to concentrated distributions. Traditional pixel-wise losses exploit strong inter-branch correlations for co-optimization, causing gradient conflicts in weakly correlated regions. Therefore, we propose an \textbf{\underline{I}}nter-\textbf{\underline{C}}hrominance and \textbf{\underline{L}}uminance inte\textbf{\underline{R}}action (ICLR) framework including a Dual-stream Interaction Enhancement Module (DIEM) and a Covariance Correction Loss (CCL). The DIEM improves the extraction of complementary information from two dimensions, fusion and enhancement, respectively. The CCL utilizes luminance residual statistics to penalize chrominance errors and balances gradient conflicts by constraining chrominance branches covariance. Experimental results on multiple datasets show that the proposed ICLR framework outperforms state-of-the-art methods.
\end{abstract}


\section{Introduction}

Low-light image enhancement (LLIE) addresses critical challenges in computer vision. While recent RGB-based methods show progress, images in low-light conditions suffer from complex degradations. Simple RGB luminance adjustments often amplify noise and distort colors \cite{guo2023low}. To mitigate this, researchers increasingly adopt color space decoupling—separating chrominance (color information) and luminance (brightness information) into distinct processing branches for independent optimization. Among decoupled spaces, the HVI color space \cite{yan2025hvi} overcomes the pure-black-plane limitation of HSV space while maintaining the one-to-one RGB mapping. This approach retains HSV’s decoupling advantages while enhancing interpretability of color transformations.

\begin{figure*}[!t]
\centering
\includegraphics[width=1.0\linewidth]{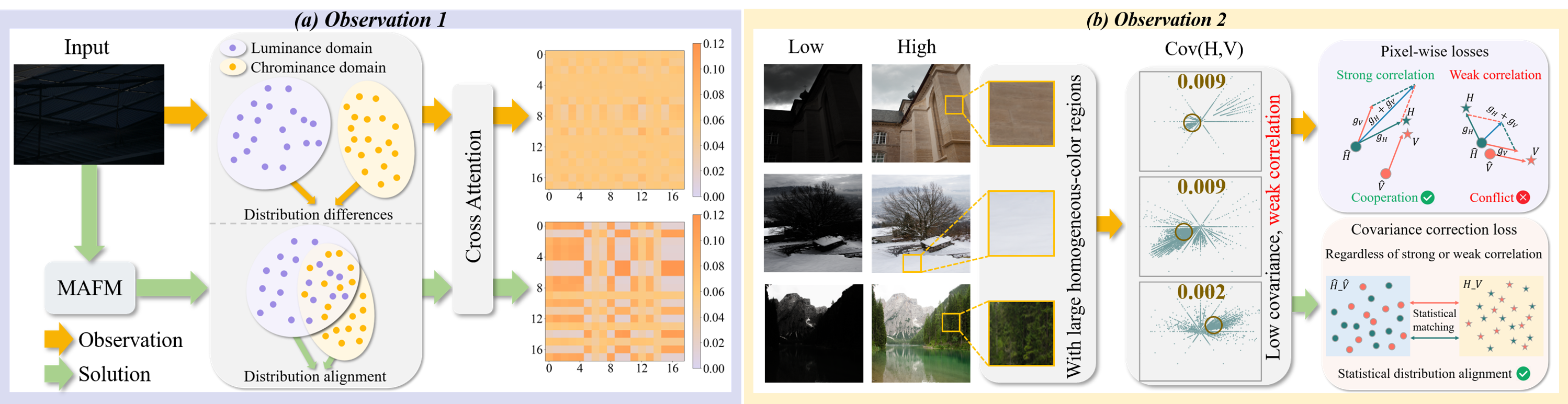}
\caption{Our two key observations. \textbf{(a)} Substantial distributional differences between luminance and chrominance limit the complementary feature extraction. \textbf{(b)} In images with large homogeneous-color regions, chrominance branches typically exhibit weak
correlation due to their relatively 
concentrated distributions, which leads to gradient conflicts.}
\label{fig:one}
\end{figure*}

CIDNet provides a new paradigm for HVI-space optimization. 
It enables luminance-chrominance interaction via cross-attention mechanisms, resolving feature separation issues in conventional decoupling methods. 
Simultaneously, it applies consistent L1 constraints to both the HVI and RGB color spaces, mitigating the separation of optimization objectives across different color spaces. 
However, 
CIDNet overlooks inherent statistical distribution relationships that exist among HVI branches, constraining network optimization. 
Based on this, we identify two novel and representative observations in natural images, supported by statistical details in the \textbf{Appendix Section B}: 

\textbf{\textit{Observation 1:}} 
As illustrated in Figure \ref{fig:one}(a), 
substantial distributional differences between luminance and chrominance cause oversmoothed cross-attention weights, limiting complementary feature extraction.
Moreover, a luminance-dependent nonlinear parameter is coupled in chrominance branches for adjusting the color-dot density in dark regions, propagating luminance errors to chrominance channels. 

\textbf{\textit{Observation 2:}} 
As illustrated in Figure \ref{fig:one}(b), images with large homogeneous-color regions 
usually exhibit weak correlation between chrominance branches due to concentrated distributions. 
However, traditional pixel-wise losses exploit strong inter-branch correlations for co-optimization, causing gradient conflicts (competing optimization directions) in weakly correlated regions. 
Consequently, conventional methods fail to optimize such areas effectively.

To address these challenges, we propose an \textbf{\underline{I}}nter-\textbf{\underline{C}}hrominance and \textbf{\underline{L}}uminance inte\textbf{\underline{R}}action (ICLR) framework. ICLR leverages complementary information between chrominance (color data) and luminance (brightness data) branches to guide dual-branch optimization while calibrating joint statistical distributions of chrominance branches.
Specifically, we improve 
the complementary information in two dimensions: enhancement and fusion. 
On the one hand, we design a Cross Dynamic Enhancement Module (CDEM) to leverage local and global contextual information to enhance complementary information. 
More importantly, we design a Multidimensional Attention-guided Fusion Module (MAFM) for chrominance branches, aiming to align chrominance and luminance branches from spatial, channel and pixel dimensions.
Concurrently, we introduce a Covariance Correction Loss (CCL) that utilizes luminance residual statistics to penalize chrominance errors, suppressing nonlinear parameter-induced diffusion errors. Meanwhile, this design balances gradient conflicts in weakly correlated distributions by constraining chrominance branches covariance. 
Our contributions can be summarized as follows:
\begin{itemize}

\item \textbf{Empirical Contribution.} We have two important observations based on the HVI: 1) Substantial distributional differences between luminance and chrominance limit complementary feature extraction, while luminance errors propagate to chrominance channels via the nonlinear parameter. 2) The chrominance branches exhibit weak correlation in images with large homogeneous-color regions, causing gradient conflicts in weakly correlated regions during optimization.

\item \textbf{Framework Contribution.} We propose an Inter-Chrominance and Luminance inteRaction (ICLR) framework, which aims to leverage complementary information between
chrominance and luminance branches to guide dual-branch optimization while calibrating joint statistical distributions of chrominance branches.

\item \textbf{Technical Contribution.} We propose a Dual-stream Interactive Enhancement Module (DIEM) and a Covariance Correction Loss (CCL). The DIEM improves the complementary information from fusion and enhancement perspectives, by incorporating Multidimensional Attention-guided Fusion Module (MAFM) and Cross Dynamic Enhancement Module (CDEM). The CCL utilizes luminance residual statistics to penalize chrominance errors and balances gradient conflicts by constraining chrominance branches covariance.

\end{itemize}

\begin{figure*}[!t]
\centering
\includegraphics[width=1.0\linewidth]{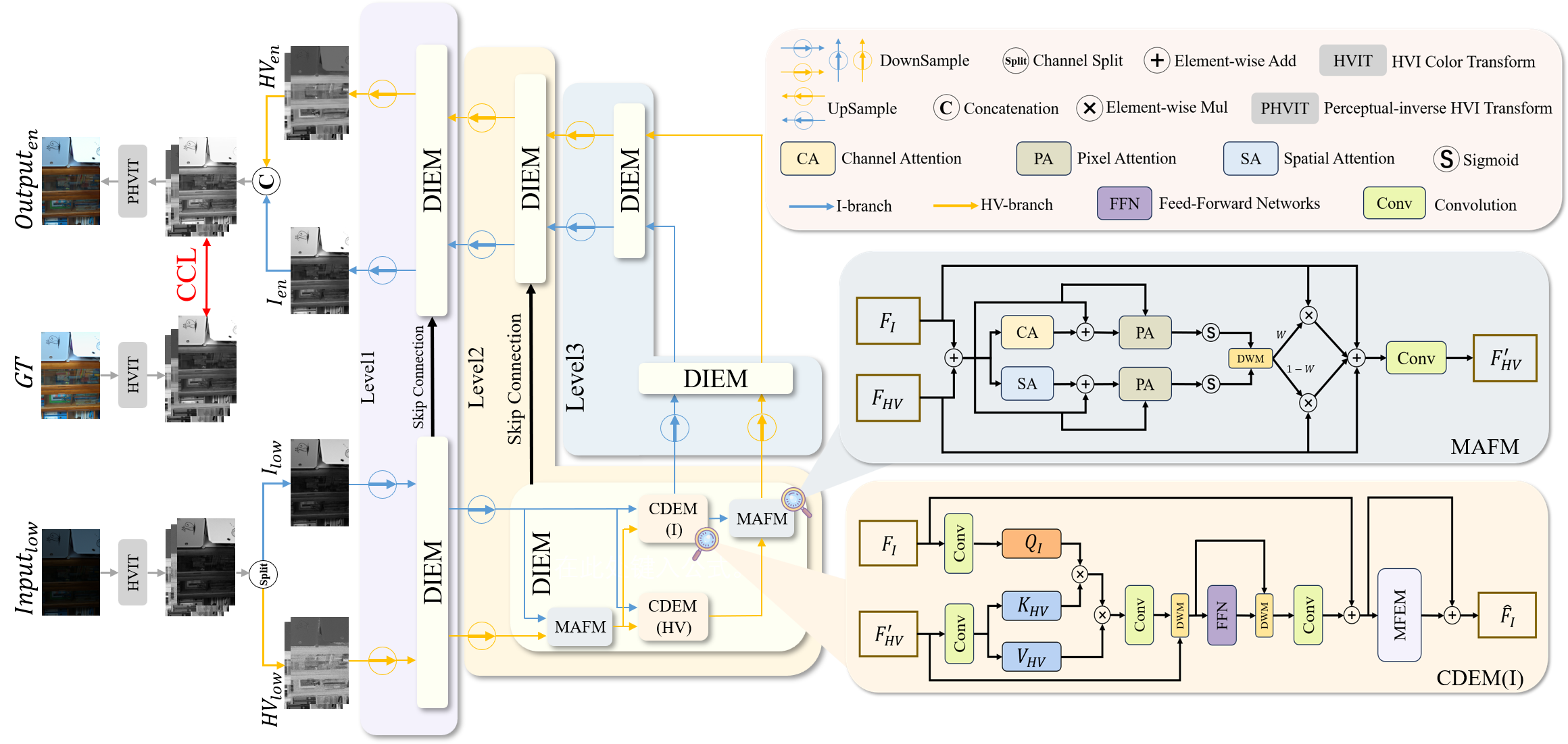}
\caption{The overall architecture of the proposed ICLR framework. ICLR is a three-level U-Net architecture integrated with Dual-stream Interaction Enhancement Modules (DIEM) and a Covariance Correction Loss (CCL). The DIEM module mainly consists of Multidimensional Attention-guided Fusion Modules (MAFM) and Cross Dynamic Enhancement Modules (CDEM).} 
\label{fig:two}
\end{figure*}

\section{Related Work}
\subsection{Low-Light Image Enhancement}
Deep learning-based methods \cite{jiang2021deep,yang2021band,jiang2022degrade} have achieved remarkable success in LLIE tasks. RetinexNet \cite{wei2018deep} decomposes low-light images into illumination and reflectance components, performing both enhancement and denoising. Zero-DCE \cite{guo2020zero} formulates LLIE as the task of estimating image-specific curves for a deep network, and introduces a zero-reference learning strategy for training. Additionally, SNRNet \cite{xu2022snr} employs a signal-to-noise ratio perceptual converter and convolutional modeling to dynamically enhance pixels through spatially varying operations. More recently, DRGN \cite{jiang2024mutual} introduces a degradation-aware two-stage generation network that first learns intrinsic degradation to simulate the distortion of environmental illumination and then refines image content to restore details and colors. Despite these advancements, however, the aforementioned deep learning methods remain confined to the study of the RGB color space.

\subsection{Color Space}
RGB channels couple both chrominance and luminance, so simple luminance enhancement adjusts all channels together, misdirecting chrominance and causing color shifts. Therefore, the paradigm of decoupling chrominance and luminance by transforming the color space is introduced to LLIE tasks. For example, HSV  \cite{chobola2024fast} and Ycbcr color space \cite{brateanu2025lyt,guo2023low}, which decouple chrominance and luminance, the conversion process with RGB color space results in multi-mapping and pure black planes, which leads to black artifacts in the image. To solve this problem, HVI color space \cite{yan2025hvi} decouples luminance while introducing Horizontal/Vertical (HV) chrominance map as a plane to quantize the chrominance reflection map, which achieves one-to-one mapping to RGB color space and eliminates the pure black plane.

\section{Methodology}
\subsection{Overall Architecture}
As illustrated in Figure \ref{fig:two}, our ICLR framework is a three-level U-Net architecture. Details of the framework can be found in \textbf{Appendix Section D.1}.

\subsection{Dual-stream Interaction Enhancement Module}
Substantial distributional differences between the two branches prevalent in natural images limit complementary feature extraction. An introduction to shortcomings of traditional cross-attention can be found in \textbf{Appendix Section D.2}. Inspired by the progressive coupling mechanism \cite{jiang2021rain}, we propose the DIEM to improve the cross-attention mechanism from dimensions of enhancement and fusion \cite{dang2024adaptive,dang2023efficient}, respectively. The core idea of DIEM is to adaptively fuse \cite{chen2024dea} the luminance and chrominance branches to increase their distributional similarity and control the fusion scale for effective complementary feature extraction, while leveraging both local and global contextual information through dedicated enhancement modules \cite{shen2024icafusion} to further enhance complementary information. Details are shown in Figure \ref{fig:two}, and here we take the enhancement of luminance branch as an example in order to clearly express the flow of the module.  

\subsubsection{Dimensions of Fusion: Multidimensional Attention-guided Fusion Module.} 
\begin{figure}[!t]
\centering
\includegraphics[width=1.0\linewidth]{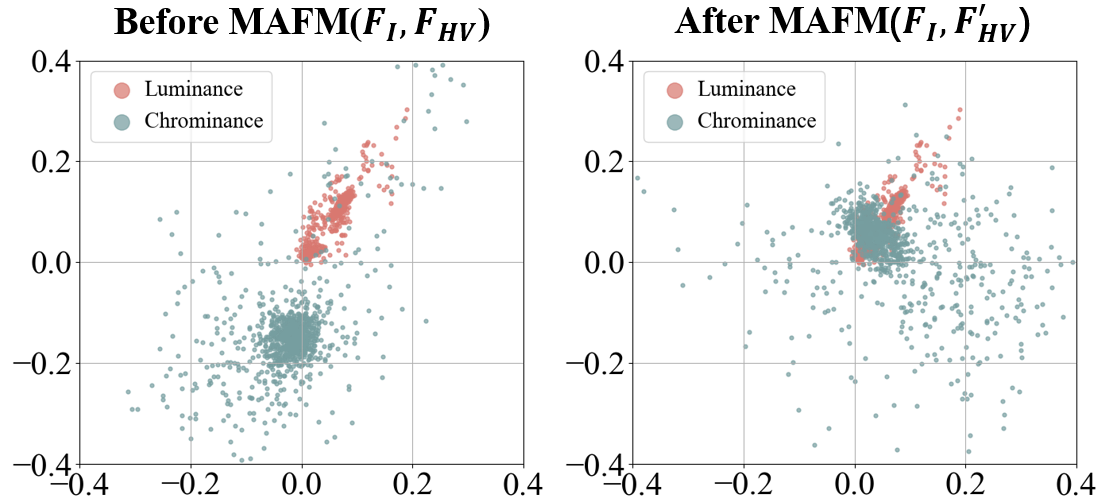}
\caption{Comparison of luminance and chrominance branches feature distributions before and after MAFM.}
\label{fig:three}
\end{figure}

There are substantial distributional differences between the chrominance and luminance branches. Inspired by adaptive fusion strategies in multi-model networks \cite{jiang2019atmfn}, we hope to effectively model the cross-subspace relationship between the two branches from different fusion dimensions to achieve adaptive distribution alignment \cite{dang2024temporo,dang2024beyond}, thereby improving the extraction of complementary information. The details of MAFM are shown in Figure \ref{fig:two}, given the input features $\mathcal{}{F}_{\mathcal{}{I}} $ and $\mathcal{}{F}_{\mathcal{}{HV}}$ of the input luminance branch and chrominance branch, the initial feature fusion is first realized to obtain the base fused features $\mathcal{}{F}_{\mathcal{}{init}}$. Then, channel attention is used to adjust weights of each channel to emphasize complementary features that are more discriminative at the global level, so that channels that are more sensitive to complementary information can dominate in subsequent fusion. Spatial attention is used in parallel to compute the mean and maximum of each spatial location, capturing the saliency of different spatial locations and ensuring that complementary information on local details is not weakened. And these two attention maps are combined with the initial fused features, allowing the attention map to reinforce the complementary information while still retaining the semantics of the original features. Then $W_c$ and $W_s$ are obtained by pixel attention by combining the original features and the two attention weights, respectively, such that pixel attention assigns an attention weight to each pixel location. It combines complementary information of chrominance and luminance at the pixel level, allowing the model to finely tune the attention at each pixel location, thus ensuring that the chrominance and luminance information is fully complementary at each spatial location. Finally, we fuse the attention weights $W_c$ and $W_s$ to obtain the final refined attention weight $W$ by dynamic weighting to ensure that the complementary features in both channel and spatial dimensions are fully integrated. And we adaptively adjust the contribution ratio of chrominance to luminance by adjusting the initial features with learned attention weights $W$ to ensure the coordination of chrominance and luminance features. We also add the original input features through residual concatenation to alleviate the gradient vanishing problem and simplify the learning process, and finally obtain the fused feature ${F_{HV}'}$. Figure \ref{fig:three} compares luminance and chrominance distributions before and after MAFM, demonstrating MAFM's capability to achieve adaptive distribution alignment. This process can be formulated as follows:
\begin{equation}
F_{init} = F_I + F_{HV},
\label{eq8}
\end{equation}
\begin{equation}
W_c = {\sigma}(PA(CA(F_{init}) + F_{init}, F_{init}),
\label{eq9}
\end{equation}
\begin{equation}
W_s = {\sigma}(PA(SA(F_{init}) + F_{init}, F_{init}),
\label{eq10}
\end{equation}
\begin{equation}
W = \varphi \cdot W_c + \omega \cdot W_s,
\label{eq11}
\end{equation}
\begin{equation}
F'_{HV} = F_{init} + W \cdot F_I + (1 - W) \cdot F_{HV},
\label{eq12}
\end{equation}
where ${\sigma}$ denotes the sigmoid operation, $CA(\cdot)$, $SA(\cdot)$ and $PA(\cdot)$ denote channel attention, spatial attention and pixel attention. Details can be found in \textbf{Appendix Section D.3}. $\varphi$ and $\omega$ represent learnable parameters.

\subsubsection{Dimensions of Enhancement: Cross Dynamic Enhancement Module.} To further enhance the representation of complementary information captured by the cross-attention mechanism, we design a novel dynamic weighting structure and a multi-branch convolutional structure from the enhancement point of view \cite{hu2023cycmunet+}, the details of which are shown in Figure \ref{fig:two}. Given the input luminance feature $\mathcal{}{F}_{\mathcal{}{I}}$ and the fused chrominance feature $F'_{HV}$, firstly, the underlying complementary information $\mathcal{}{Z}_{\mathcal{}{I}}$ is captured through the cross-attention mechanism, and then the dynamic weighting mechanism is utilized to regulate the priority enhancement of the complementary information. On the one hand, the structured cues captured in the luminance branch are forced to directly intervene in the reconstruction of chrominance features to uplift the complementary information in the region of strong cross-subspace correlation into the dominant restoration signals, while on the other hand, the original luminance features can be effectively retained to strengthen the expression of complementary information. Next, the nonlinear mapping of feed-forward neural network is utilized to fully explore the correlation between luminance and chrominance in high-dimensional space, and the reconstruction of the feature flow is further adaptively adjusted to obtain the feature $\mathcal{}\hat{Z}_{\mathcal{}{I}}$ through the dynamic weighting mechanism. Finally, the MFEM utilizes the multi-branch convolutional structure to extract features in different directions, scales, and semantic spaces in parallel, thereby enhancing the ability to express complementary information, to obtain the final enhanced feature $\mathcal{}\hat{F}_{\mathcal{}{I}}$. This process can be formulated as follows:
\begin{equation}
Z_{I} = {CrossAttention}(F_{I}, F'_{HV}),
\label{eq13}
\end{equation}
\begin{equation}
\hat{Z}_{I} = \lambda \cdot FFN(\alpha \cdot Z_{I} + \beta \cdot F'_{HV}) + \mu \cdot Z_{I} \\,
\label{eq14}
\end{equation}
\begin{equation}
\hat{F}_{I} = \mathcal{F}_{MFEM}(F_{I} + \hat{Z}_{I}) + (F_{I} + \hat{Z}_{I}),
\label{eq15}
\end{equation}
where ${CrossAttention(\cdot)}$ denotes the cross-attention mechanism, ${FFN(\cdot)}$ stands for feed-forward neural network in standard Transformer. $\mathcal{F}_{MFEM}(\cdot)$ represents the multi-branch feature enhancement module. $\alpha$, $\beta$,  $\lambda$ and $\mu$ represent the learnable parameters.

\begin{figure*}[!t]
\centering
\includegraphics[width=1.0\linewidth]{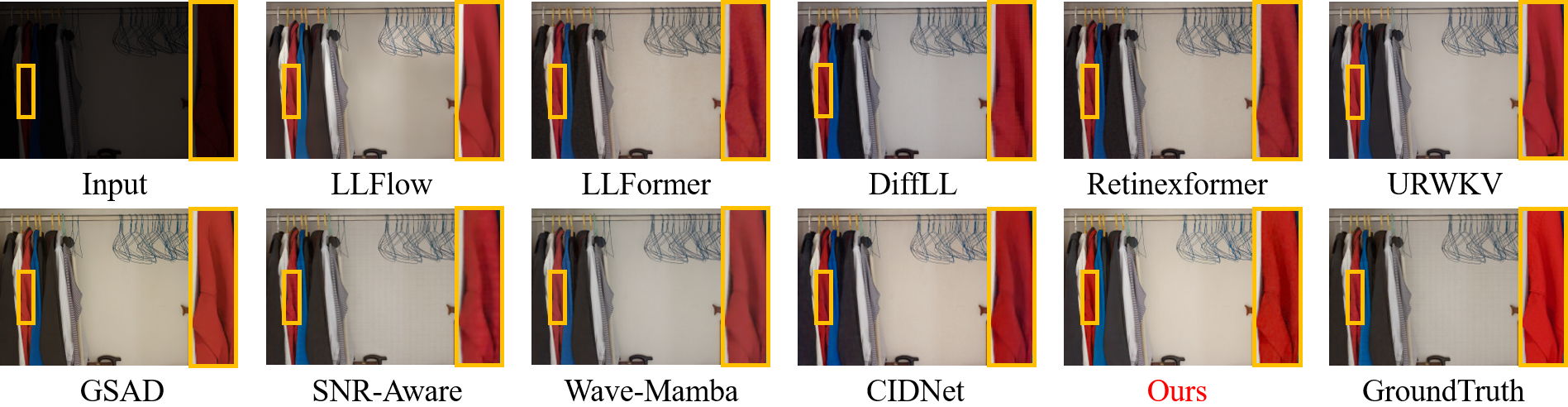}
\caption{The qualitative comparison of the methods we compared in terms of color restoration. Zoom in for the best view.}
\label{fig:four}
\end{figure*}

\subsection{Covariance Correction Loss (CCL)}
\subsubsection{Error Correction for Luminance Guidance.} In the HVI, the nonlinear parameter $C_k$ is used to adjust the color point density of the low-luminance color plane thus allowing it to be coupled in chrominance branches. This parameter can be considered as a function of the luminance $C_k(I)$, where $I\in[0,1]$ and $\frac{\pi \hat{I}}{2} \in \left[0, \frac{\pi}{2} \right]$. So the function $C_k(I)$ is a monotonically increasing function on the range $[0,1]$. For luminance error $\Delta I =\hat{I} - I$, when the luminance error $\Delta I$ is increasing, the error in the nonlinear parameter $\Delta C_k$ is also increased. Since the nonlinear parameters are coupled in the calculation of the chrominance branch, the corresponding chrominance channels errors $\Delta H$ and $\Delta V$ also increase. Therefore, the luminance error is diffused to the chrominance channels through the nonlinear parameter $C_k$, which makes the prediction of the chrominance branch heavily shifted. To address this problem, we propose to guide the correction of chrominance branches errors through luminance, and more specifically, we use the mean and variance of the luminance error as adaptive weights to adjust the optimization of the chrominance branches, and aggravate the penalty on the chrominance branches errors when the luminance error is large in order to mitigate the effect of the luminance error diffused by the nonlinear parameter $C_k$:
\begin{equation}
L_Z = \frac{1}{N} \sum_{i=1}^{N} (\hat{Z}_i - Z_i)^2, \quad Z \in \{I, H, V\}
\label{eq16}
\end{equation}
\begin{equation}
W_H = 1.0 + \frac{1}{N} \sum_{i=1}^{N} |\hat{I}_i - I_i|,
\label{eq17}
\end{equation}
\begin{equation}
W_V = 1.0 + \sqrt{\frac{1}{N} \sum_{i=1}^{N} (|\hat{I}_i - I_i| - \frac{1}{N} \sum_{i=1}^{N} |\hat{I}_i - I_i|)^2 },
\label{eq18}
\end{equation}
\begin{equation}
L_{I-HV} = W_HL_H+W_VL_V,
\label{eq19}
\end{equation}
where $N=B\times H\times W$ represents the total number of pixels per batch, $L_H$, $L_V$ and $L_I$ denote the independent losses of the three channels of HVI, respectively. $W_H$ and $W_V$ stand for the adaptive weight of the chrominance branches penalty strength. $L_{I-HV}$ represents the luminance-guided error correction loss function.

\subsubsection{Covariance Statistical Constraints.} Conventional loss minimize the discrepancy between the predicted image and the ground truth by pixel-wise constraints, and these constraints have a tendency to be present in the HVI: stronger linear correlation between chrominance branches $H$ and $V$ correlates with higher prediction quality, while weaker correlation leads to degradation. This phenomenon arises because when $H$ and $V$ are strongly linearly correlated, their gradients update directions align, enabling simultaneous optimization of both components through a unified parameter adjustment, and when $H$ and $V$ are weakly linearly correlated, their gradients require updates in divergent or even opposing directions, leading to conflicts in parameter optimization and cancellation effects that undermine model convergence. To mitigate gradient competition caused by weak linear correlation between chrominance branches $H$ and $V$, we shifts the focus from a deterministic pixel-wise comparison to a statistical perspective, emphasizing the learning of distributions rather than individual pixel values \cite{dangdiff,danghallucination}. The core idea is to optimize the overall pixel distribution by means of a loss function corresponding to the computation of the fitted covariance, which leads to the optimization of the overall pixel distribution and gets rid of the reiance on the pixel-wise linear correlation between $H$ and $V$. More specifically, the formula for calculating the covariance between $H$ and $V$ can be expressed as: 
\begin{equation}
Cov(H,V)=E(HV)-E(H)E(V),
\label{eq20}
\end{equation}
where $Cov(\cdot)$ denotes the covariance function, $E(\cdot)$ denotes the mean value function.

\begin{table*}[]
\centering
\setlength{\tabcolsep}{1.2mm}
\begin{tabular}{c|c|ccc|ccc|ccc|c}
\hline
\multirow{2}{*}{Methods} & \multicolumn{1}{l|}{\multirow{2}{*}{References}} & \multicolumn{3}{c|}{LOLv1}                        & \multicolumn{3}{c|}{LOLv2-Real}                                                     & \multicolumn{3}{c|}{LOLv2-Syn}                                                      & \multicolumn{1}{l}{\multirow{2}{*}{Params(M)}} \\
                         & \multicolumn{1}{l|}{}                            & PSNR↑           & SSIM↑          & LPIPS↓         & \multicolumn{1}{l}{PSNR↑} & \multicolumn{1}{l}{SSIM↑} & \multicolumn{1}{l|}{LPIPS↓} & \multicolumn{1}{l}{PSNR↑} & \multicolumn{1}{l}{SSIM↑} & \multicolumn{1}{l|}{LPIPS↓} & \multicolumn{1}{l}{}                           \\ \hline
LLFlow                   & AAAI'22                                          & 24.046          & 0.860          & 0.136          & 26.428                    & 0.903                     & 0.096                       & 19.219                    & 0.824                     & 0.214                       & 37.68                                          \\
SNR-Aware                & CVPR'22                                          & 24.609          & 0.842          & 0.105          & 30.923                    & 0.893                     & 0.092                       & 16.251                    & 0.767                     & 0.192                       & 39.13                                          \\
Retinexformer            & ICCV'23                                          & 25.155          & 0.845          & 0.085          & 28.983                    & 0.882                     & 0.068                       & 16.190                    & 0.771                     & \textbf{0.189}                       & 1.53                                           \\
LLFormer                 & AAAI'23                                          & 23.649          & 0.816          & 0.169          & 27.749                    & 0.860                     & 0.143                       & 17.163                    & 0.784                     & 0.244                       & 24.55                                          \\
GSAD                     & NeurIPS'23                                       & 27.595          & 0.875          & 0.091          & 28.818                    & 0.895                     & 0.095                       & 19.784                    & 0.806                     & 0.225                       & 17.36                                          \\
DiffLL                   & TOG'23                                           & 24.634          & 0.798          & 0.201          & 27.781                    & 0.845                     & 0.180                       & 19.269                    & 0.766                     & 0.279                       & 22.07                                          \\
Wave-Mamba               & MM'24                                            & 25.847          & 0.858          & 0.143          & 30.714                    & 0.905                     & 0.103                       & 19.700                    & 0.819                     & 0.233                       & \textbf{1.51}                                  \\
URWKV                    & CVPR'25                                          & 26.513          & 0.869          & 0.107          & 31.413                    & 0.906                     & 0.073                       & 20.518                    & 0.810                     & 0.214                       & 2.25                                           \\
CIDNet                   & CVPR'25                                          & 27.732          & 0.870          & 0.117          & 31.436                    & 0.896                     & 0.101                       & 20.375                    & 0.816                     & 0.243                       & 1.88                                           \\
Ours                     & -                                                & \textbf{28.603} & \textbf{0.885} & \textbf{0.068} & \textbf{32.320}           & \textbf{0.907}            & \textbf{0.054}              & \textbf{20.719}           & \textbf{0.837}            & 0.217                       & 4.27                                           \\ \hline
\end{tabular}
\caption{Quantitative results on LOLv1, LOLv2-Real and LOLv2-Syn. The best performance is marked in bold. To compare the generalization performance of methods, We train on the LOLv1 and test on LOLv1, LOLv2-Real and LOLv2-Syn.}
\label{tab:one}
\end{table*}

The independent channel losses for $H$ and $V$ in Eq.\ref{eq16}, $L_H$ and $L_V$ have already constrained the individual means $E(H)$ and $E(V)$. To further enforce covariance-based constraints on the global statistical distribution of $H$ and $V$ pixels, we must additionally constrain the joint mean term $E(HV)$, and this constraint can be formulated as follows:
\begin{equation}
L_{HV} = \frac{1}{B} \sum_{i=1}^{B} (\frac{1}{HW} \sum_{i=1}^{H} \sum_{j=1}^{W} \hat{H}_i \hat{V}_j - \frac{1}{HW} \sum_{i=1}^{H} \sum_{j=1}^{W} H_i V_j)^2,
\label{eq21}
\end{equation}
where $B$, $H$ and $W$ denote batch size, height and width, respectively, $L_{HV}$ represents the joint mean-constrained loss.

Finally, we simple combine $L_I$, $L_{I-HV}$ and $L_{HV}$ these three losses as the final objective CCL.

\section{Experiments}

\subsection{Datasets and metrics}
We train and evaluate our model on the LLIE benchmark dataset LOL, which consists of two versions: LOLv1 \cite{wei2018deep} and LOLv2 \cite{yang2021sparse}. LOLv2 is divided into two subsets: LOLv2-real and LOLv2-synthetic. The training and testing splits are $485$:$15$ for LOLv1, $689$:$100$ for LOLv2-real and $900$:$100$ for LOLv2-synthetic. In addition, we test on five unpaired datasets, DICM \cite{lee2013contrast}, LIME \cite{guo2016lime}, MEF \cite{ma2015perceptual}, NPE \cite{wang2013naturalness} and VV \cite{vonikakis2018evaluation}, to validate generalization performance. For paired datasets, we adopt Peak Signal-to-Noise Ratio (PSNR) \cite{wang2004image}, Structure Similarity Index Measure (SSIM) \cite{wang2004image}, and Learned Perceptual Image Patch Similarity (LPIPS) \cite{zhang2018unreasonable} as evaluation metrics. For unpaired datasets, we adopt Natural Image Quality Evaluator (NIQE) \cite{mittal2012making} as evaluation metric. Among them, PSNR and SSIM are better when higher, while LPIPS and NIQE are better when lower. Experiments with more datasets can be found in \textbf{Appendix Section E.1}, which further validate the performance of our method on large-scale datasets.

\subsection{Experimental settings}
We set the patch size to $256\times256$, the batch size to $8$, and train the model with the Adam optimizer for a total of $1000$ epochs on a single NVIDIA $3090$ GPU by using PyTorch. The learning rate is initially set to $1\times10^{-4}$ and then steadily decreased to $1\times10^{-7}$ by the cosine annealing scheme during the training process. We train on the LOLv1 dataset and test on the LOLv1, LOLv2-real and LOLv2-synthetic datasets to demonstrate the generalization performance of our method.

\subsection{Comparisons with State-of-the-Art Methods}
We compare our method with state-of-the-art (SOTA) methods for LLIE, including LLFlow \cite{wang2022low}, SNR-Aware \cite{xu2022snr}, Retinexformer \cite{cai2023retinexformer}, LLFormer \cite{wang2023ultra}, GSAD \cite{hou2023global}, DiffLL \cite{jiang2023low}, Wave-Mamba \cite{zou2024wavemamba}, URWKV \cite{xu2025urwkv}, and CIDNet \cite{yan2025hvi}. 

\begin{figure}[!t]
\centering
\includegraphics[width=1.0\linewidth]{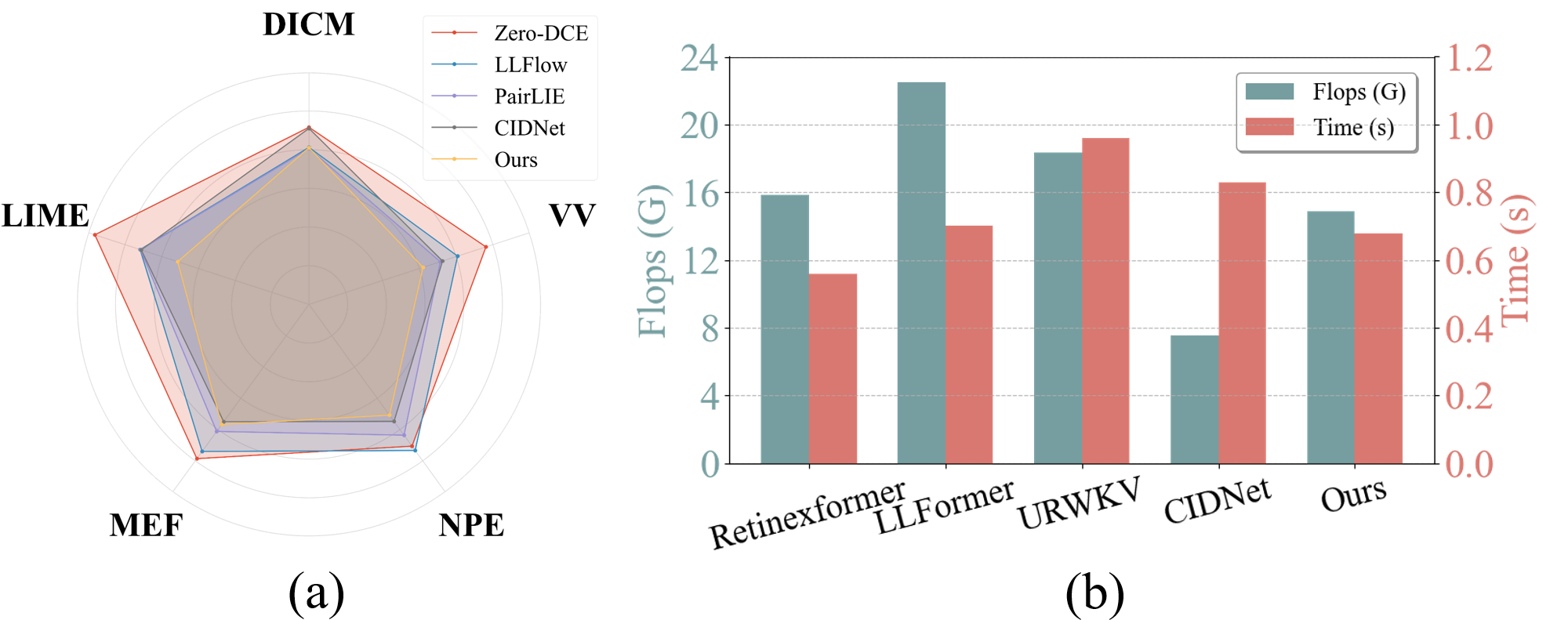}
\caption{\textbf{(a)} Comparison of NIQE↓ on five unpaired datasets, where lower values indicate better performance. \textbf{(b)} Comparison of Flops and Inference Time.} 
\label{fig:five}
\end{figure}

\subsubsection{Comparison on LOLv1 and LOLv2 Datasets.} 
\begin{figure*}[!t]
\centering
\includegraphics[width=1.0\linewidth]{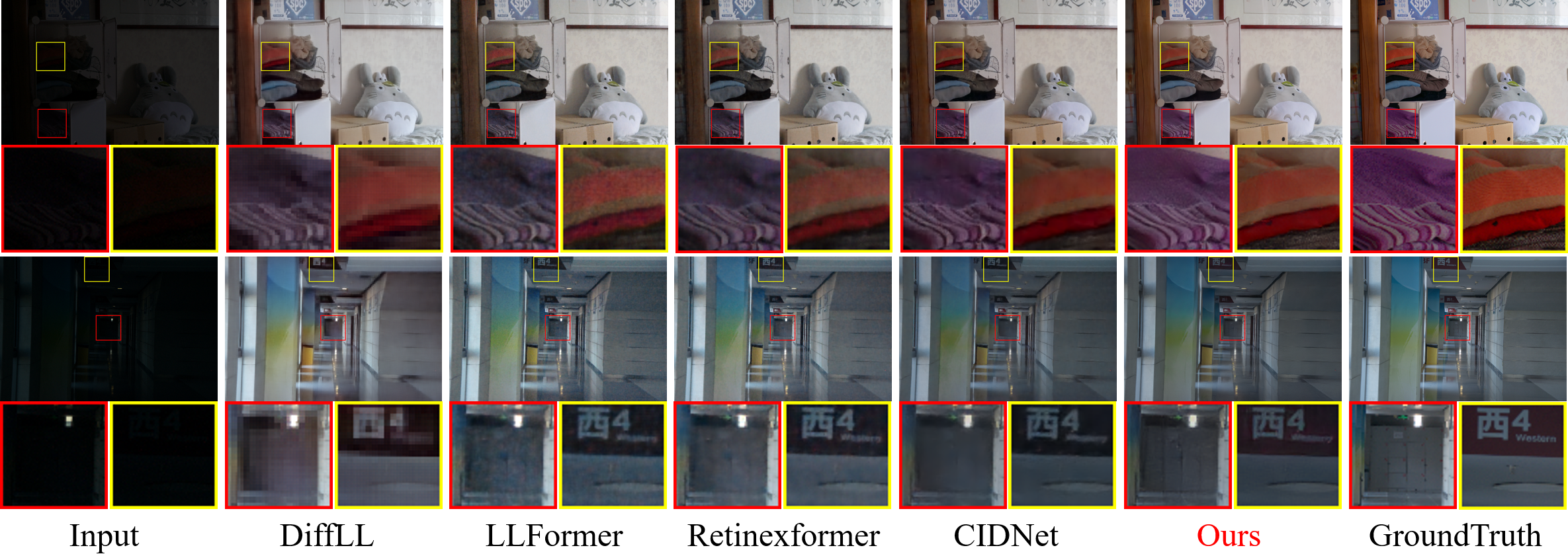}
\caption{Visual comparisons of the enhanced results by different methods on LOLv1 and LOLv2. Zoom in for the best view.}
\label{fig:seven}
\end{figure*}

Table \ref{tab:one}, Figure \ref{fig:four} and Figure \ref{fig:seven} list quantitative and qualitative comparisons with SOTA methods on  LOLv1 and LOLv2. For quantitative comparison, our method gains $0.87$ dB, $0.88$ dB and $0.20$ dB on LOLv1, LOLv2-real and LOLv2-syn datasets, respectively. This indicates that our method has better generalization performance, mainly because we enhanced the expression of complementary information between the chrominance and luminance branches through the dual-stream interaction enhancement module, enabling the capture of multi-scale features more effectively for better generalization and adaptability across datasets. For qualitative comparison, our method is able to restore more faithful colors than other methods for the purple and orange sweaters in the first row and the red sign in the second row. Moreover, the red clothing in Figure \ref{fig:four} demonstrates that our method can restore a more vibrant red color compared to other methods. The advantage of our method in color restoration is attributed to the CCL function specifically designed for chrominance branches. More visual comparison experiments can be found in \textbf{Appendix
Section E.3}.

\subsubsection{Comparison on unpaired datasets.} Figure \ref{fig:five}(a) shows a quantitative comparison of the NIQE metric on five unpaired datasets. It can be observed that our method achieves the lowest values on all four datasets except for a slightly higher value than CIDNet on the MEF dataset, demonstrating the superior generalization capability of our method.

\subsubsection{Comparison on computational overhead.} Table \ref{tab:one} and Figure \ref{fig:five}(b) show comparisons of parameters, flops and inference time. It can be observed that although our method has higher parameters than URWKV, we achieve advantages in both flops and inference time. This indicates that our method can achieve a noticeable improvement in restoration quality at the cost of small computational overhead.

\begin{figure}[!t]
\centering
\includegraphics[width=1.0\linewidth]{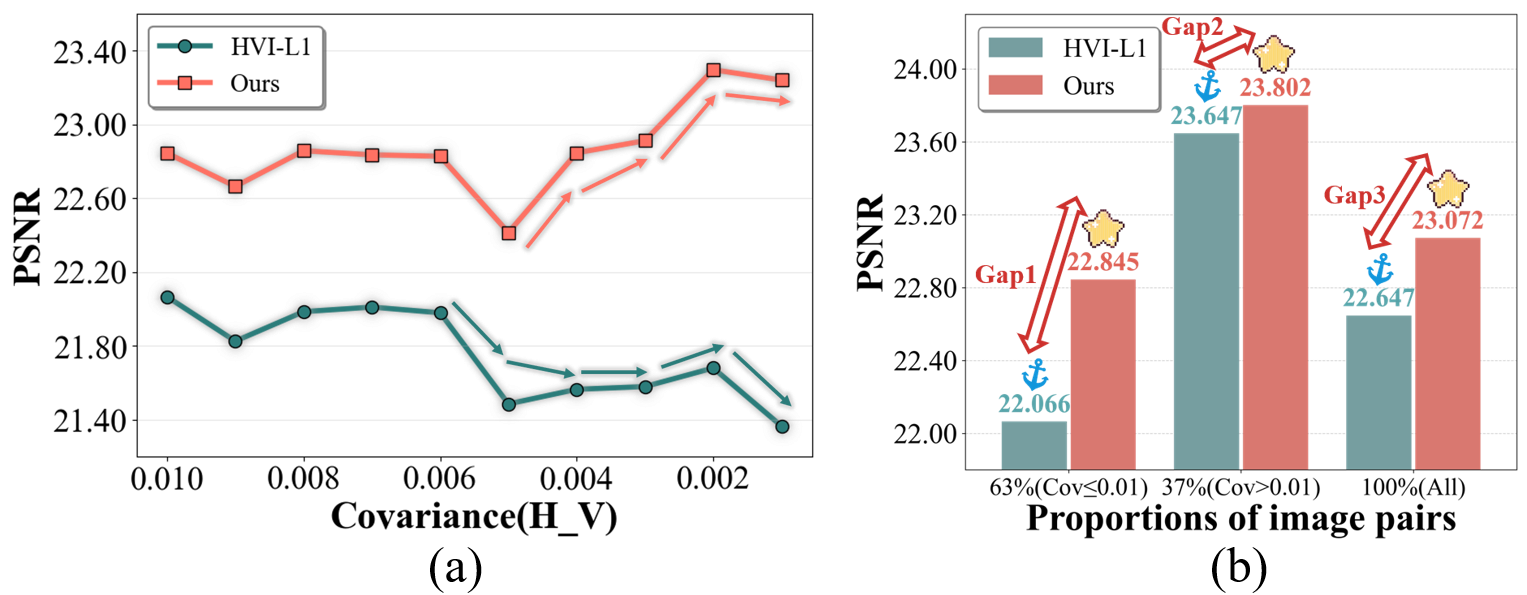}
\caption{Evaluation and comparison of our method and conventional pixel-wise loss under different covariance conditions, using PSNR to evaluate performance.}
\label{fig:six}
\end{figure}

\subsubsection{Comparison on a New Synthetic Dataset.} 
To further demonstrate the superiority of our method over pixel-wise losses approaches in resolving gradient conflicts, we merged the test sets from LOLv1, LOLv2-real and LOLv2-synthetic into a new dataset containing a total of $215$ low-light image pairs. The new synthetic dataset consists of $215$ images, of which $\textbf{136}$ \textbf{images} ($\textbf{63\%}$) have a covariance coefficient less than or equal to 0.01, while $\textbf{79}$ \textbf{images} ($\textbf{37\%}$) have a covariance coefficient greater than $0.01$. As shown in Figure \ref{fig:six}(a), for image pairs with covariances less than or equal to $0.01$, we divided the covariance range into more detailed categories and evaluated the PSNR. It can be clearly observed that the pixel-wise loss shows a significant downward trend in PSNR as the covariance decreases. In contrast, our method not only outperforms the pixel-wise loss under all covariance conditions. Moreover, this advantage becomes even more pronounced when the covariance is relatively low. To illustrate the specific mechanism of our method for low-covariance images, we evaluated images with proportions of $63\%$ ($Cov\leq0.01$), $37\%$ ($Cov>0.01$) and 100\% ($\mathrm{All}$), as shown in Figure \ref{fig:six}(b). Through quantitative analysis, $90\%$ of the overall performance improvement (Gap$3$) comes from improvements in the portion with a covariance less than or equal to $0.01$ (Gap$1$), while the remaining $10\%$ comes from improvements in the portion greater than $0.01$ (Gap$2$). This result shows that our method can effectively solve the gradient conflict problem caused by pixel-wise loss under low-covariance conditions.

\begin{table}[H]
\centering
\setlength{\tabcolsep}{0.5mm}
\begin{tabular}{c|ccc|ccc}
\hline
\multirow{2}{*}{Sets} &      & Components &            &        & Metrics & \multicolumn{1}{l}{}       \\
                         & MAFM & CDEM       & CCL        & PSNR↑  & SSIM↑   & \multicolumn{1}{l}{LPIPS↓} \\ \hline
$\Omega1$                 & \ding{52}    & \ding{52}          & L1         & 28.202 & 0.879   & 0.073                      \\
$\Omega2$                 & \ding{52}    & \ding{52}          & L2         & 28.275 & 0.880   & 0.073                      \\
$\Omega3$                 & \ding{52}    & TCA        & \ding{52}          & 27.998 & 0.883   & 0.074                      \\
$\Omega4$                 & \ding{56}    & \ding{52}          & \ding{52} & 28.028 & 0.880   & 0.080                      \\
$\Omega5$                 & \ding{52}    & \ding{52}          & \ding{52}          & 28.603 & 0.885   & 0.068                      \\ \hline
\end{tabular}
\caption{Ablation Studies on MAFM, CDEM and CCL.}
\label{tab:two}
\end{table}

\subsection{Ablation Study}
In this section, we conduct ablation studies on the main components of our ICLR framework using the LOLv1 dataset. 

\subsubsection{Effectiveness of MAFM.} As shown in Table \ref{tab:two}, settings for $\Omega4$ and $\Omega5$ indicate that the PSNR without MAFM decreased by $0.575$ dB. Combined with the distribution comparison before and after MAFM in Figure \ref{fig:three}, this indicates that MAFM aligns chrominance and luminance branches to improve complementary feature extraction.

\subsubsection{Effectiveness of CDEM.} We compare the advantages of CDEM over Traditional Cross Attention (TCA) in two experimental settings, $\Omega3$ and $\Omega5$. The PSNR decreased by $0.605$ dB when CDEM is replaced by TCA, demonstrating that CDEM more effectively leverages local and global contextual information to enhance complementary features.

\begin{figure}[H]
\centering
\includegraphics[width=1.0\linewidth]{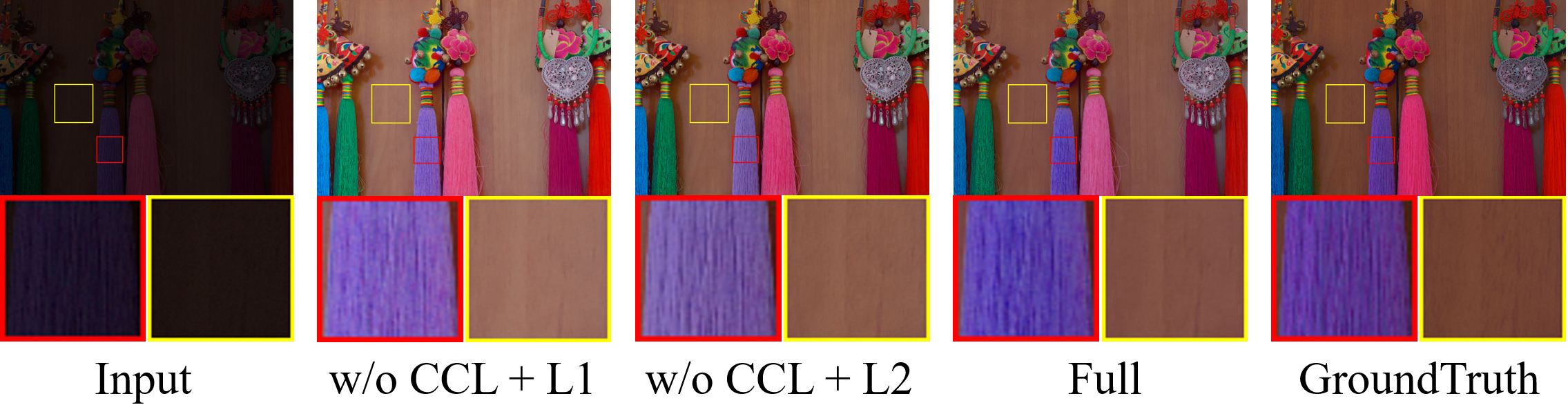}
\caption{Visual comparison of CCL with conventional pixel-wise losses. Please zoom in for the best view.}
\label{fig:eight}
\end{figure}

\subsubsection{Effectiveness of CCL.} To compare CCL with conventional pixel-wise losses for color restoration, We replace CCL with L1 and L2 in settings $\Omega1$ and $\Omega2$. Compared with setting $\Omega5$, the PSNR decreased by $0.401$ dB and $0.328$ dB. Combined with the visual comparison in Figure \ref{fig:eight}, this demonstrates that CCL can more effectively optimize the chrominance branches through covariance constraints, leading to a more faithful restoration of color information.

\subsubsection{Effectiveness of the number of MAFMs.} As shown in Table \ref{tab:three}. We further conduct ablation studies on the two MAFMs before and after the DIEM. Compared to having both MAFM($1$) (before) and MAFM($2$) (after), having only MAFM($1$) or only MAFM($2$) resulted in PSNR decreases of $0.231$ dB and $0.268$ dB, respectively. This indicates that both MAFMs play an important role in aligning the chrominance and luminance branches.

\begin{table}[H]
\centering
\setlength{\tabcolsep}{1.0mm}
\begin{tabular}{c|cc|ccc}
\hline
\multirow{2}{*}{Sets} & \multicolumn{2}{c|}{Components} &        & Metrics & \multicolumn{1}{l}{} \\
                      & MAFM(1)        & MAFM(2)        & PSNR↑  & SSIM↑   & LPIPS↓               \\ \hline
$\Omega1$                    & \ding{56}              & \ding{56}              & 28.028 & 0.880   & 0.080                \\
$\Omega2$                    & \ding{56}              & \ding{52}              & 28.372 & 0.883   & 0.073                \\
$\Omega3$                    & \ding{52}              & \ding{56}              & 28.335 & 0.882   & 0.072                \\
$\Omega4$                    & \ding{52}              & \ding{52}              & 28.603 & 0.885   & 0.068                \\ \hline
\end{tabular}
\caption{Ablation Studies on the number of MAFMs.}
\label{tab:three}
\end{table}

\section{Conclusion}
In this paper, in order to solve the problems of interaction between chrominance and luminance branches, as well as within chrominance branches for natural images, we propose an Inter-Chrominance and Luminance inteRaction (ICLR) framework. Specifically, we improve the complementary feature extraction in the interaction of luminance and chrominance branches through a Multidimensional Attention-guided Fusion Module (MAFM) and a Cross Dynamic Enhancement Module (CDEM) in terms of the dimensions of fusion and enhancement, respectively. In addition, we design a Covariance Correction Loss (CCL) to optimize the interaction between chrominance branches from a statistical distribution perspective. 

\section{Acknowledgements}
This research was financially supported by the National Natural Science Foundation of China (62501189, 62376201), Hubei Provincial Science \& Technology Talent Enterprise Services Program (2025DJB059), Hubei Provincial Special Fund for CentralGuided Local S\&T Development (2025CSA017), and the Natural Science Foundation of Heilongjiang Province of China for Excellent Youth Project (YQ2024F006).

\bibliography{main}

\clearpage
\end{document}